\documentclass[11pt,a4paper]{article}
\pdfoutput=1
\usepackage[hyperref]{acl2018}
\usepackage{times}
\usepackage{latexsym}
\usepackage{bm}
\usepackage{url}
\usepackage{hyperref}

\aclfinalcopy 
\def\aclpaperid{1180} 

\title{Party Matters: Enhancing Legislative Embeddings with Author Attributes for Vote Prediction }

\author{Anastassia Kornilova \qquad Daniel Argyle \qquad Vlad Eidelman \\
 FiscalNote \\
  {\tt \{anastassia.kornilova,daniel,vlad\}@fiscalnote.com} 
  }

\date{}

\begin{document}
\maketitle
\begin{abstract}
Predicting how Congressional legislators will vote is important for understanding their past and future behavior. However, previous work on roll-call prediction has been limited to single session settings, thus did not consider generalization across sessions. 
In this paper, we show that metadata is crucial for modeling voting outcomes in new contexts, as changes between sessions lead to changes in the underlying data generation process. 
We show how augmenting bill text with the sponsors' ideologies in a neural network model can achieve an average of a $4\%$ boost in accuracy over the previous state-of-the-art.
\end{abstract}

\section{Introduction}
\label{sec:intro}
Quantitative analysis of the voting behavior of legislators has long been a problem of interest in political science, and recently in NLP as well~\cite{gerrish,kraft}. 
One of the most popular techniques in political science for modeling legislator behavior is the application of spatial, or ideal point, models built from voting records~\cite{poole,clinton}, that are often used to represent uni-dimensional or multi-dimensional ideological stances. While {\it roll call votes} (i.e Congressional voting records) provide explanatory power about a legislators position with respect to previously voted-on bills, these models are limited to in-sample analysis, and are thus incapable of predicting votes on new bills.



%
%
%
%
%
%
To address this limitation, recent work has introduced methods that take advantage the text of the bill, along with the voting records, to model Congressional voting behavior 
\citep{gerrish,nguyen:15, kraft}. This work is related to a long line of studies on using political text to model behavior, ranging over political books, Supreme Court decisions, speeches and Twitter \citep{mosteller1963inference, getout, idealspeech, scotus, idealbook, president, idealtwitter}.

In addition to enabling prediction, associating text with ideology allows for a further degree of interpretability. However, all previous work incorporating text into roll call prediction have limited their evaluation to in-session training and testing.\footnote{A session is a 2-year period of legislative business.}

As legislators typically serve for multiple sessions, and similar bills are proposed across sessions, we want to be able to leverage this data across sessions to inform our model.
However, the generalizability of previous methods to a cross-session setting is unknown. 

In this work, we explore the problem of roll call prediction across sessions. We show that previous methods are unable to generalize across sessions, thus suggesting that current text representations are not sufficient for modeling voting outcomes in new contexts. We hypothesize that each session has a different underlying data generation process, wherein the ideological position of the observed bills varies depending on the controlling party. This is supported by the observation that about $75\%$ of bills up for a vote in a given session have a sponsor in the party in power. 

As noted in \citet{text_reuse}, the policy area, or topic, of the bill, and the ideological position, are two separate dimensions underlying the text. Since legislators tend to sponsor bills that are ideologically aligned with them, a model trained on a single session will mostly be exposed to bills with a specific ideology on each topic. Thus, a single session model may get the ideology information as an implicit prior without needing to explicitly capture it. This challenge was not obvious in previous studies that were limited to a single session. Across sessions, however, the ideological prior on a given topic changes, resulting in variations in voting patterns that are not captured by current text modeling methodologies alone. 

In applications where the text may contain an insufficient signal, researchers may turn to additional metadata features. This technique has previously been used in various contexts, such as incorporating sponsor and committee features for predicting bill committee survival \cite{yano2012textual}, and enhancing tweet recommendations with location data \cite{meta}. 

We propose a neural architecture that directly models the ideological variation across sessions using metadata about the bill sponsors, and show that this can strongly improve performance with little overhead to complexity and training time.

 \section{Model}
 \label{sec:model}
Spatial voting models assume that a legislator has a numeric {\it ideal point} which represents their ideology. They make voting decisions on bills, which also have a numeric representation. While the details of the implementation vary,\footnote{For example, Poole and Rosenthal represent bills as cutpoints that divide legislators into yes and no groups \cite{poole} and later work based on item response theory conceptualizes bills as "discrimination" vectors that are mutiplied by an ideal point vector.} spatial voting models share the idea that the closer a bill's representation is to a legislator's ideal point the more likely the legislator is to vote {\it yes}.

Following this framework, we model the core vote prediction problem as follows: Given a legislator, $L$, and a bill, $B$, predict their vote $y$, with possible outcomes: {\it yes} or {\it no}. 

Using these inputs, let $v_L$ be an embedding representing the legislator, and $v_B$ be the bill embedding. 
First, $v_B$ is projected into the legislator embedding space: \begin{equation} v_{BL} = \mathbf{W_B} v_B + \mathbf{b_B}\end{equation}
where $\mathbf{W_B}$ and $\mathbf{b_B}$ are a weight matrix and a bias vector, respectively. Then, we measure the alignment between the two vectors. Previous work used a dot-product for this step, instead, we express the comparison as follows:
\begin{equation} \mathbf{W_v} (v_{BL} \odot v_L) + \mathbf{b_v} \end{equation}
where $\odot$ represents element-wise multiplication, and $\mathbf{W_v}$ is a weight vector of the same dimensions as $v_L$. Finally, we apply a sigmoid activation function to get the vote prediction:
\begin{equation} p(y=yes| B, L) = \sigma(\mathbf{W_v} (v_{BL} \odot v_L) + \mathbf{b_v})\end{equation}

Using this architecture, we develop several novel bill representations. First, we consider different text-only representations, then we show how to incorporate metadata.

\subsection{Text Model}
\label{sec:text_model}
Previous work incorporating text has primarily been based on  topic models~\cite{gerrish, Lauderdale:14, nguyen:15} and embeddings~\cite{kraft}. As the embedding framework achieved superior performance, we adopt a similar architecture.
While~\citet{kraft} represented the text using a mean word embedding (MWE) representation, we replace it with a Convolutional Neural Network (CNN) representation~\cite{kim}, which has achieved superior performance on recent text classification tasks~\cite{goodcnn:3, goodcnn:2,goodcnn:1}. Our CNN uses 4-grams and 400 filter maps. 

%
%
%
%

\subsection{Sponsor Metadata}
\label{sec:spon_meta}
We posit that a legislator's voting behavior is influenced both by the topic and the ideology of a bill. A legislator may be more liberal on one issue and more conservative on another. Thus, we need to capture both aspects. While previous work has shown that text alone contains ideological information~\cite{Iyyer:2014}, the metadata of the bill may be a stronger source, especially for ideology. This approach has had success in the related problem of bill committee survival,\footnote{Congressional bills, first, are voted on in a committee, before moving to the floor.}
where signals about the sponsors, committee and chamber were used in conjunction with text models \cite{yano2012textual}. 

We use this idea to improve our bill representations. One particularly strong signal is the author of the bill, because of their ideological motives. For simplicity, we represent the bill's authorship as the percentage of Republican and Democrat sponsors ($p_r$ and $p_d$).  We propose that the Republican and Democratic sponsors influence the text of the bill in different ways. To obtain the overall ideological position of the bill, we combine the versions of the bill influenced by each party. The final bill can thus be represented as follows:
\begin{equation} v_B = ((\mathbf a_r p_r) \cdot T_r) + ((\mathbf a_d p_d) \cdot T_d) \end{equation}
where $T_r$ and $T_d$ are the Republican and Democratic copies of the text representation (e.g MWE or CNN); $p_r$ and $p_d$ are the scalars representing the percentage of sponsors from each party (e.g 0.7 and 0.3); and $\mathbf {a_r}$ and $\mathbf {a_p}$ are vectors representing how the percentages should influence each dimension of the text embedding.

The larger $p_r$ or $p_d$ is, the stronger the influence of that party on the bill. 

We test two text representations for $T_r$ and $T_d$: one using MWEs and one using CNNs. The underlying word embeddings are initialized with 50d GloVE vectors~\cite{glove} and are non-static during training. 

The rest of the model weights are initialized randomly with the {\it glorot uniform} distribution \cite{glorot}. The length of $v_L$ is set to 25. All models are trained using binary cross-entropy loss and optimized with the {\it AdaMax} algorithm \cite{adamax}. The models are trained for 50 epochs, using mini-batches of size 50. 
%
%

\section{Dataset}
\label{sec:data}
 Our dataset was collected from GovTrack,\footnote{\url{https://theunitedstates.io/}} and consists of nonunanimous roll call votes and texts of resolutions and bills introduced in the 106th to 111th Congressional sessions.\footnote{We exclude bills with unanimous votes because these are typically associated with routine matters (for example, the naming a post office or an official commendation) that do not contain ideological motivation.  We consider bills where less than 1\% of legislators voted `no' to be unanimous; about 42\% of bills fall into this category.}
We also collect the bill summaries written by the Congressional Research Service\footnote{\url{ https://www.congress.gov/help/legislative-glossary/}} (a non-partisan organization), that provide shorter descriptions of the key actions in each bill. All text is preprocessed by lowercasing and removing stop-words.

As bills are often much longer than the typical document encountered in other NLP tasks, with an average of 2683 words per bill, and some bills having hundreds of pages, with correspondingly lengthy summaries, this poses a problem for our compositional neural architecture. 
To address this, we limit the length of each full-text and summary to $N$ words, where $N$ is empirically set to the 80$^{th}$ percentile of the collection. For summaries $N$=400, and for full-text $N$=2000.
\section{Experiments}
\label{sec:exp}
As described earlier, the experimental framework in previous work treated each session individually. 
To evaluate the ability of our model to generalize across sessions, we perform several sets of experiments. In the first set, in-session, we perform 5 fold cross-validation over the 2005-2012 sessions. In the second, out-of-session, we train on multiple sessions, 2005-2012, and evaluate on sessions not included during training, the 2013-2014 and 2015-2016 sessions. During testing, we only include legislators present in the training data.

The overall statistics for our dataset are presented in Tables~\ref{tbl:data_counts} and~\ref{tbl:party}.
  
\begin{table}[]
\centering
\label{my-table2}
\begin{tabular}{|c|c|c|c|}
\hline
{\color[HTML]{000000} }
Session & Total Bills  & Total Votes & \multicolumn{1}{l|}{\begin{tabular}[c]{@{}l@{}}\% Yes\\ Votes\end{tabular}}   \\ \hline

2005-2012 & 1718 &  685,091 & 68.4\% \\ 
2013-2014 & 360 &  136,807 & 66.4\% \\ 
2015-2016  & 382 & 153,605 & 61.8\% \\ \hline
\end{tabular}
\caption{Count of Bills and Votes  \label{tbl:data_counts}}
\end{table}

\begin{table}[]
\centering
\label{my-label}
\begin{tabular}{|c|c|c|}
\hline
{\color[HTML]{000000} }
Session & House Majority  & Senate Majority  \\ \hline
2005-2006 & R &  R \\ 
2007-2008 & D &  D \\ 
2009-2010 & D &  D \\ 
2011-2012 & R &  D \\ 
2013-2014 & R &  D \\ 
2015-2016  & R & R \\ \hline
\end{tabular}
\caption{Party in power by session  \label{tbl:party}}
\end{table}

\section{Results}
\label{sec:results}


To understand how sponsor parties and text interact in the input, and how our predictive power changes when testing on in-session bills and out-of-session bills. We test the following models:

\begin{itemize}
\item MWE: mean word embedding text model as described in \citet{kraft} using summaries;
\item MWE+FT: MWE model using full bill text;
\item CNN: text model from Section~\ref{sec:text_model} over summaries;
\item MWE+Meta: MWE representation combined with metadata as described in Section~\ref{sec:spon_meta};
\item CNN+Meta: like MWE+Meta but using a CNN instead of averaging;
\item MWE+Meta+FT: As above using full bill text;
\item Meta-Only: A variation on MWE+Meta that uses the same, random ``dummy'' text for all the bills, only changing the metadata ($p_r$ and $p_d$).
\end{itemize}

Each model is first evaluated in-session, where both train and test bills come from the same set of sessions, and thus same distribution, and then out-of-session, where training bills are from one set of sessions and the model is evaluated on a different set. All results are presented in Table~\ref{tbl:results}.

\subsection{In-session Results}
We evaluate our models with accuracy on 5-fold cross-validation. All three models combining text with metadata perform significantly better than the others, showing that the text and meta information have complimentary predictive power, and that our models' sponsor-augmented text representation is able to capture the ideological preference. 
The CNN+Meta achieves the highest accuracy of $86.21$, followed by MWE+Meta at $85.96$, showing that the CNN learns a somewhat better text representation than MWE. Compare this to the baseline MWE model without meta information, which achieves an accuracy of $81.10$, only slightly better than the Meta-Only model at $80.27$. Contrary to our hypothesis, MWE achieves higher accuracy than Meta-Only. However, it remains unclear whether this signal is related to ideology or other contextual information. The performance on the out-of-session setting will determine whether this signal is akin to ideology.

\subsection{Out-of-session Results}
In this setting, on both test sessions, text with meta information achieves the best performance as well. 
On the 2013-2014 session, the CNN+Meta model does the best at $83.59$. 
Unlike the in-session setting, Meta-only does better than the text-only models (MWE, CNN). This supports the theory that within the sessions we are able to capture contextual ideology from the text, but once we move to a new session the text models no longer contain an accurate representation of the Congressional ideology.

While in other experiments we are able to achieve at least a $17\%$ improvement over the Guess Yes baseline, on 2015-2016, the best model, MWE+Meta, is only able to achieve a $10\%$ gain. During this session divisions arose within the Republican party in the House of Representatives that disrupted the typical voting dynamics.\footnote{A conservative bloc of the Republican Party (the “Freedom Caucus”) began to assert influence over party leadership, eventually resulting in the ouster of John Boehner as Speaker \cite{freedom}. } Unlike 2013-2014, the Meta-Only model does worse than the text ones; however, the gap between them is much smaller. 

\begin{table}[]
\centering
\begin{tabular}{l|c||cc|} \hline
{} & \multicolumn{1}{c||}{ in-session }&  \multicolumn{2}{c|} {out-of-session} \\ \hline
{     }  & \multicolumn{1}{l||}{\begin{tabular}[c]{c@{}} 2005-\\ 2012\end{tabular}} & \multicolumn{1}{l|}{\begin{tabular}[c]{@{}l@{}}2013-\\ 2014\end{tabular}} & \multicolumn{1}{l|}{\begin{tabular}[c]{@{}l@{}}2015-\\ 2016\end{tabular}} \\ \hline
         \multicolumn{1}{l|}{Guess Yes}  & 68.31   & 65.92  & 61.07  \\ \hline
\multicolumn{1}{l|}{MWE}        & 81.10  & 77.57 & 69.80  \\ \hline
\multicolumn{1}{l|}{MWE + FT} &    81.46 & 68.33  & 57.94 \\ \hline
\multicolumn{1}{l|}{CNN}        & 83.24  & 77.49 & 69.63  \\ \hline
\multicolumn{1}{l|}{Meta-Only}  & 80.87  & 82.28 & 67.10  \\ \hline
\multicolumn{1}{l|}{MWE + Meta} & 85.96  & 82.73 & \textbf{71.90} \\ \hline
\multicolumn{1}{l|}{MWE+Meta+FT} & 85.14 &  82.43  & 69.86 \\ \hline
\multicolumn{1}{l|}{CNN + Meta} & \textbf{86.21} & \textbf{83.59}  & 70.99   \\ \hline
 \end{tabular}
\caption{Accuracy Results\label{tbl:results}}
\end{table}

\subsection{Overall Analysis}  
These experiments provide several interesting insights.
First, because using both text and metadata (MWE+Meta or CNN+Meta) results in the strongest model in every case, we confirm that legislators vote based on both the topic and the ideology of the bill. 

Second, the text-only models do significantly worse on the out-of-session tests than the in-session ones. This confirms our theory that session-specific contextual information is implicitly captured by the previous single-session models, but that context is not accurate in new sessions. If we were capturing ideology from the text, then the text only model should have performed well out-of-session. 

Third, to further examine whether a neural model was the best technique for modeling text with metadata, we trained a SVM model over the bag-of-words representation of the summary, indicator variables for the legislators and the percent of bill sponsors in each party (e.g $p_d$). This model did not perform as well as either MWE or Meta-Only, showing that the embedding approach is better at representing this combination of features.

Finally, the models that embed the full text (+FT) generally perform worse than embedding the summaries. While this confirms that the summary contains sufficient information about the topics and the actions in the bill, we did not fully explore the bill text. 

\section{Future Work}
While Congress introduces close to $20,000$ bills every session, very few of them receive a vote, limiting the dataset. We would like to explore various bootstrapping techniques that would allow us to expand the dataset size with artificial votes.

Furthermore, while our text representations are sufficient for modeling shorter text, i.e. summaries, we would like to test more sophisticated representations in the future, in particular, those designed to handle longer texts.
\section{Conclusion}
In this paper, we developed a neural network architecture to predict legislators votes that augments bill text with sponsor metadata. We introduced a new evaluation setting for this task: out-of-session performance; which allows us to examine the generalizability of our proposed model, and was not considered in past studies. Finally, we showed that the introduction of metadata to bias the text representations outperforms the existing text-based methods in all experimental settings. 
\bibliographystyle{acl_natbib}
\bibliography{acl2018}

\begin{thebibliography}{25}
\expandafter\ifx\csname natexlab\endcsname\relax\def\natexlab#1{#1}\fi

\bibitem[{Clinton et~al.(2004)Clinton, Jackman, and Rivers}]{clinton}
Joshua Clinton, Simon Jackman, and Douglas Rivers. 2004.
\newblock The statistical analysis of roll call data.
\newblock \emph{American Political Science Review}, 98(2).

\bibitem[{Dauphin et~al.(2016)Dauphin, Fan, Auli, and Grangier}]{goodcnn:3}
Yann~N. Dauphin, Angela Fan, Michael Auli, and David Grangier. 2016.
\newblock \href {http://arxiv.org/abs/1612.08083} {Language modeling with gated
  convolutional networks}.
\newblock \emph{CoRR}, abs/1612.08083.

\bibitem[{Gerrish and Blei(2011)}]{gerrish}
Sean Gerrish and David~M Blei. 2011.
\newblock Predicting legislative roll calls from text.
\newblock In \emph{Proceedings of ICML}.

\bibitem[{Glorot and Bengio(2010)}]{glorot}
Xavier Glorot and Yoshua Bengio. 2010.
\newblock Understanding the difficulty of training deep feedforward neural
  networks.
\newblock In \emph{Proceedings of AISTATS}.

\bibitem[{Iyyer et~al.(2014{\natexlab{a}})Iyyer, Enns, Boyd-Graber, and
  Resnik}]{idealbook}
Mohit Iyyer, Peter Enns, Jordan Boyd-Graber, and Philip Resnik.
  2014{\natexlab{a}}.
\newblock Political ideology detection using recursive neural networks.
\newblock In \emph{Proceedings of ACL}, volume~1, pages 1113--1122.

\bibitem[{Iyyer et~al.(2014{\natexlab{b}})Iyyer, Enns, Boyd-Graber, and
  Resnik}]{Iyyer:2014}
Mohit Iyyer, Peter Enns, Jordan Boyd-Graber, and Philip Resnik.
  2014{\natexlab{b}}.
\newblock \href {docs/2014_acl_rnn_ideology.pdf} {Political ideology detection
  using recursive neural networks}.
\newblock In \emph{Association for Computational Linguistics}.

\bibitem[{Kim(2014)}]{kim}
Yoon Kim. 2014.
\newblock Convolutional neural networks for sentence classification.
\newblock \emph{arXiv:1408.5882}.

\bibitem[{Kingma and Ba(2014)}]{adamax}
Diederik~P. Kingma and Jimmy Ba. 2014.
\newblock \href {http://arxiv.org/abs/1412.6980} {Adam: {A} method for
  stochastic optimization}.
\newblock \emph{CoRR}, abs/1412.6980.

\bibitem[{Kraft et~al.(2016)Kraft, Jain, and Rush}]{kraft}
Peter Kraft, Hirsh Jain, and Alexander~M Rush. 2016.
\newblock An embedding model for predicting roll-call votes.
\newblock In \emph{Proceedings of EMNLP}.

\bibitem[{Lauderdale and Clark(2014)}]{Lauderdale:14}
Benjamin~E. Lauderdale and Tom~S. Clark. 2014.
\newblock Scaling politically meaningful dimensions using texts and votes.
\newblock \emph{American Journal of Political Science}, 58(3):754--771.

\bibitem[{Linder et~al.(2018)Linder, Desmarais, Burgess, and
  Giraudy}]{text_reuse}
Fridolin Linder, Bruce~A. Desmarais, Matthew Burgess, and Eugenia Giraudy.
  2018.
\newblock Text as policy: Measuring policy similarity through bill text reuse.
\newblock \emph{SSRN/2812607}.

\bibitem[{Lizza(2017)}]{freedom}
Ryan Lizza. 2017.
\newblock \href {https://www.newyorker.com/magazine/2015/12/14/a-house-divided}
  {The war inside the republican party}.

\bibitem[{Mosteller and Wallace(1963)}]{mosteller1963inference}
Frederick Mosteller and David~L Wallace. 1963.
\newblock Inference in an authorship problem: A comparative study of
  discrimination methods applied to the authorship of the disputed federalist
  papers.
\newblock \emph{Journal of the American Statistical Association},
  58(302):275--309.

\bibitem[{Nguyen et~al.(2015)Nguyen, Boyd{-}Graber, Resnik, and
  Miler}]{nguyen:15}
Viet{-}An Nguyen, Jordan~L. Boyd{-}Graber, Philip Resnik, and Kristina Miler.
  2015.
\newblock \href {http://aclweb.org/anthology/P/P15/P15-1139.pdf} {Tea party in
  the house: {A} hierarchical ideal point topic model and its application to
  republican legislators in the 112th congress}.
\newblock In \emph{Proceedings of ACL}.

\bibitem[{Pennington et~al.(2014)Pennington, Socher, and Manning}]{glove}
Jeffrey Pennington, Richard Socher, and Christopher Manning. 2014.
\newblock Glove: Global vectors for word representation.
\newblock In \emph{Proceedings of EMNLP}, pages 1532--1543.

\bibitem[{Poole and Rosenthal(1985)}]{poole}
Keith~T Poole and Howard Rosenthal. 1985.
\newblock A spatial model for legislative roll call analysis.
\newblock \emph{American Journal of Political Science}, pages 357--384.

\bibitem[{Preo{\c{t}}iuc-Pietro et~al.(2017)Preo{\c{t}}iuc-Pietro, Liu,
  Hopkins, and Ungar}]{idealtwitter}
Daniel Preo{\c{t}}iuc-Pietro, Ye~Liu, Daniel Hopkins, and Lyle Ungar. 2017.
\newblock Beyond binary labels: political ideology prediction of twitter users.
\newblock In \emph{Proceedings of ACL}.

\bibitem[{Sim et~al.(2013)Sim, Acree, Gross, and Smith}]{president}
Yanchuan Sim, Brice~DL Acree, Justin~H Gross, and Noah~A Smith. 2013.
\newblock Measuring ideological proportions in political speeches.
\newblock In \emph{Proceedings of EMNLP}.

\bibitem[{Sim et~al.(2016)Sim, Routledge, and Smith}]{scotus}
Yanchuan Sim, Bryan~R. Routledge, and Noah~A. Smith. 2016.
\newblock Friends with motives: Using text to infer influence on scotus.
\newblock In \emph{Proceedings of EMNLP}.

\bibitem[{Thomas et~al.(2006)Thomas, Pang, and Lee}]{getout}
Matt Thomas, Bo~Pang, and Lillian Lee. 2006.
\newblock \href {http://arxiv.org/abs/cs/0607062} {Get out the vote:
  Determining support or opposition from congressional floor-debate
  transcripts}.
\newblock \emph{CoRR}, abs/cs/0607062.

\bibitem[{Wen et~al.(2016)Wen, Zhang, Luo, and Wang}]{goodcnn:2}
Ying Wen, Weinan Zhang, Rui Luo, and Jun Wang. 2016.
\newblock \href {http://arxiv.org/abs/1606.06905} {Learning text representation
  using recurrent convolutional neural network with highway layers}.
\newblock \emph{CoRR}, abs/1606.06905.

\bibitem[{Xing and Paul(2017)}]{meta}
Linzi Xing and Michael~J Paul. 2017.
\newblock Incorporating metadata into content-based user embeddings.
\newblock In \emph{Proceedings of the 3rd Workshop on Noisy User-generated
  Text}, pages 45--49.

\bibitem[{Yang et~al.(2016)Yang, MacDonald, and Ounis}]{goodcnn:1}
Xiao Yang, Craig MacDonald, and Iadh Ounis. 2016.
\newblock \href {http://arxiv.org/abs/1606.07006} {Using word embeddings in
  twitter election classification}.
\newblock \emph{CoRR}, abs/1606.07006.

\bibitem[{Yano et~al.(2012)Yano, Smith, and Wilkerson}]{yano2012textual}
Tae Yano, Noah~A Smith, and John~D Wilkerson. 2012.
\newblock Textual predictors of bill survival in congressional committees.
\newblock In \emph{Proceedings of NAACL}. Association for Computational
  Linguistics.

\bibitem[{Yu et~al.(2008)Yu, Kaufmann, and Diermeier}]{idealspeech}
Bei Yu, Stefan Kaufmann, and Daniel Diermeier. 2008.
\newblock Classifying party affiliation from political speech.
\newblock \emph{Journal of Information Technology \& Politics}, 5(1):33--48.

\end{thebibliography}

\end{document}